# EuSQuAD: Automatically Translated and Aligned SQuAD2.0 for Basque

## *EuSQuAD: SQuAD2.0 Traducido y Alineado Automáticamente para Euskera*

**Aitor García-Pablos,[1*] Naiara Perez,[2*] Montse Cuadros,[1] Jaione Bengoetxea[2]**
[1]Vicomtech Foundation, Basque Research and Technology Alliance (BRTA)
[2]HiTZ Center - Ixa, University of the Basque Country (UPV/EHU)
{agarciap,mcuadros}@vicomtech.org

**Abstract:** The widespread availability of Question Answering (QA) datasets in English has greatly facilitated the advancement of the Natural Language Processing (NLP) field. However, the scarcity of such resources for minority languages, such as Basque, poses a substantial challenge for these communities. In this context, the translation and alignment of existing QA datasets plays a crucial role in narrowing this technological gap. This work presents EuSQuAD, the first initiative dedicated to automatically translating and aligning SQuAD2.0 into Basque, resulting in more than 142k QA examples. We demonstrate EuSQuAD's value through extensive qualitative analysis and QA experiments supported with EuSQuAD as training data. These experiments are evaluated with a new human-annotated dataset.

**Resumen:** La amplia disponibilidad de conjuntos de datos de preguntas y respuestas en inglés ha facilitado en gran medida el avance del campo de Procesamiento de Lenguaje Natural (PLN). Sin embargo, la escasez de tales recursos para idiomas minoritarios, como el euskera, plantea un desafío sustancial para estas comunidades. En este contexto, la traducción y alineación de conjuntos de datos desempeña un papel crucial en la reducción de esta brecha tecnológica. Este trabajo presenta EuSQuAD, la primera iniciativa dedicada a traducir y alinear automáticamente SQuAD2.0 al euskera. Demostramos el valor de EuSQuAD a través de un extenso análisis cualitativo y experimentos de QA, para los cuales se ha creado además un nuevo dataset anotado por humanos.

## 1 Introduction

Question Answering (QA) is a core machine reading comprehension task in the field of Natural Language Processing (NLP). In its most classical form, extractive QA, systems are presented with a question and a text passage or *context* in which they must identify the corresponding answer span.

The popularity of QA has only grown over the years (Zeng et al., 2020) fuelled by its diverse range of applications, and recent advancements in generative NLP have further spurred interest in the field. For instance, researchers are now leveraging these datasets as crucial resources for instructing generative models (Wei et al., 2022; Wang and others, 2022, among others), enabling them to generalise effectively across different tasks.

The availability of datasets like SQuAD2.0 (Rajpurkar, Jia, and Liang, 2018) has been pivotal in driving these advancements. At the same time, the fact that these resources are limited to hegemonic languages like English only serves to widen the preexisting gap with respect to research communities working with minority languages, such as Basque.

In an effort to enhance access and availability to annotated data in this language, we present EuSQuAD[1], the first Basque version of SQuAD2.0 and the largest QA dataset in Basque to date, with over 142k question-answer pairs grounded in 20k passages.

EuSQuAD has been generated automatically by *1)* machine-translating SQuAD2.0 to

[*] Equal contribution.

[1] `https://github.com/Vicomtech/EusQuAD`

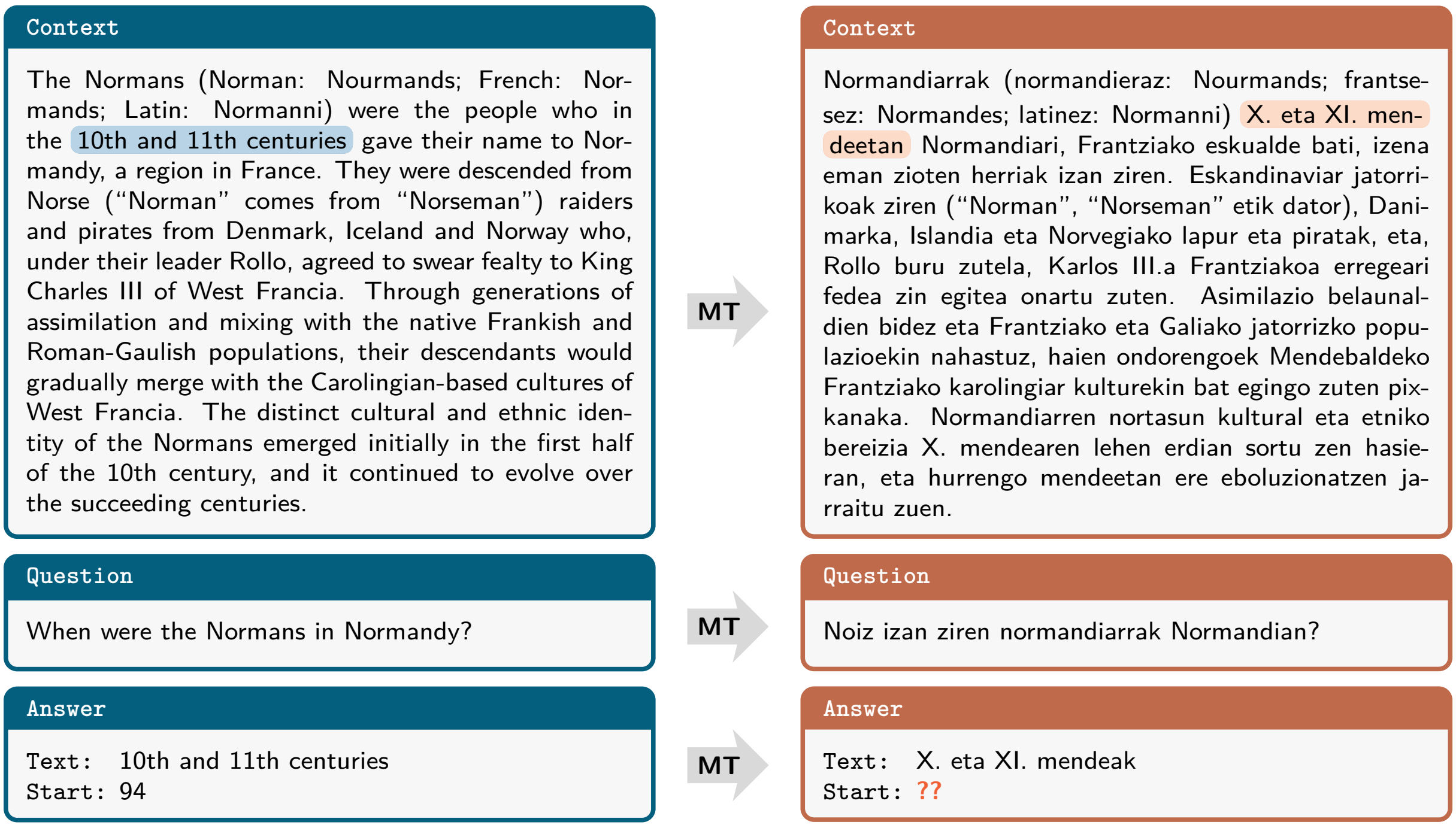

Figure 1: A SQuAD2.0 instance machine-translated from English (left) to Basque (right), where the Basque instance is corrupted due to the answer not being a substring of the context.

Basque; then, *2)* aligning context and answer spans, i.e., rectifying corrupted translated instances (see Figure 1) through semantic text similarity based on neural language models.A manual analysis of EuSQuAD reveals error rates lower than those of previous attempts to automatically translate SQuAD to other languages. In addition, we further demonstrate EuSQuAD's value by evaluating multiple state-of-the-art models fine-tuned on EuSQuAD against an original QA dataset in Basque of 490 questions, manually annotated for this purpose.

Interestingly, the results obtained suggest that embeddings produced by character-based language models are better suited for alignment purposes —within the parameters of our investigation— than the more widespread token-based ones. Despite our focus on the Basque language, we hypothesise that this finding is likely to apply to other languages as well, particularly to agglutinative languages like Basque.

In summary, this work makes the following contributions: *(a)* We release EuSQuAD, a high-quality synthetic dataset of extractive QA in Basque for system training purposes, comprising more than 142k instances. *(b)* We present an additional QA dataset of 490 questions, manually annotated for testing purposes. *(c)* We conduct a series of experiments that evidence EuSQuAD's value and provide insights into the suitability of character- and token-based language models for span alignment purposes.

The rest of the article is structured as follows. Section 2 provides an overview of the related work, with a special emphasis on Basque QA and works that automate the translation of SQuAD. Sections 3 and 4 describe, respectively, the process followed to obtain EuSQuAD and the manually labelled test set. In Section 5, we report and analyse the result of the above mentioned experiments. Section 6 discusses the limitations of this work and suggests future lines of work. Section 7 concludes the article by summarising the main conclusions of the study.

## *2 Related work*

**QA in Basque** The landscape of question answering (QA) in the Basque language has been relatively limited, with only a few notable endeavours to date. One significant effort was undertaken during the CLEF 2008 international workshop (Forner and others, 2009), with tasks on both monolingual and cross-lingual QA in several languages, including Basque. The challenge consisted in answering various sets of questions drawing upon a pool of documents, each set focusing on a specific topic. This dataset is not available to the public. A more recent contribution to the field is ElkarHizketak (Otegi et al., 2020). Akin to the approach taken by QuAC (Choi et al., 2018), this study diverged from traditional QA paradigms by focusing on conversational interactions, thereby introducing nuances distinct from those encountered in conventional QA scenarios. Despite these contributions, a crucial gap remains in the domain of extractive QA. To date, no publicly available dataset exists for this fundamental task in Basque. Our study addresses this gap by releasing EuSQuAD, a new dataset tailored specifically for extractive QA in Basque.

**QA in other languages** SQuAD2.0 (Rajpurkar, Jia, and Liang, 2018) is the latest iteration of the SQuAD dataset (Rajpurkar et al., 2016). It combines the original 100k questions with an additional set of over 50k *unanswerable* or *impossible* questions, which are questions that cannot be answered based on the provided context. The SQuAD dataset family is widely recognised and extensively used in the field of extractive QA, supporting a popular community leaderboard[2] with held out test sets. While most of the datasets similar to SQuAD have been developed in English —e.g., Trivia QA (Joshi et al., 2017), Natural Questions (Kwiatkowski and others, 2019)—, the research community has put significant effort into creating QA datasets in other languages too. For instance, the French Question Answering Dataset (FQuAD) consists of more than 60k samples manually crafted from the French Wikipedia (d'Hoffschmidt et al., 2020). Snæbjarnarson and Einarsson (2022) released more than 18k crowd-sourced Natural Questions in Icelandic. For Slovak, there is SK-QuAD (Hládek et al., 2023), a manually annotated QA dataset of more than 91k questions, including unanswerable questions and plausible answers. XQuAD (Artetxe, Ruder, and Yogatama, 2020) is a cross-lingual benchmark that comprises 240 paragraphs and 1,190 question-answer pairs from SQuAD v1.1 professionally translated into ten languages. See the survey by Chandra et al. (2021) for more references on QA data in languages other than English.

[2] `rajpurkar.github.io/SQuAD-explorer/`

**Automated SQuAD translations** Developing quality resources such as those mentioned above is a costly process, particularly when dealing with the creation of tens of thousands of question-answer pairs required to approximate the performance of English models. For this reason, several works have focused on leveraging existing datasets like SQuAD by machine-translating them into the desired target language. This approach requires the availability of a reliable translation model for the relevant language pair, and a strategy to locate the answers in the translated context. In this vein, Tasmiah Tahsin Mayeesha and Rahman (2021) translated SQuAD2.0 into Bengali and used the Levenshtein distance between translated answers and translated context spans to identify the correct answer span. While this method reportedly rectified the majority of the corrupted answers, it is unable to handle cases where the answer was translated with a synonymous expression. Carrino, Costa-jussà, and Fonollosa (2020) circumvent this problem with the their "Translate Align Retrieve" approach. The authors translate SQuADv1.1 contexts and questions to Spanish —but not answers; instead, they retrieve the correct answer span from the translated context through a token alignment map between source and target contexts. According to the authors, this method yielded moderate results, with 50% of the resulting question-answer pairs being incorrect upon manual error analysis. Works like Mozannar et al. (2019) for Arabic and Abadani et al. (2021) for Persian choose to simply discard faulty answers or retain them unchanged. In this work, we build on Tasmiah Tahsin Mayeesha and Rahman (2021) but propose employing state-of-the-art neural language models to compute the semantic similarity between the translated answer and context spans, thus decreasing alignment errors to 23%.

## 3 *EuSQuAD*

EuSQuAD is the first large-scale, synthetic extractive QA dataset for Basque. It is the translation of SQuAD2.0's training and development data, the two publicly available partitions. To evaluate our approach, we introduce a new manually annotated test set for Basque QA in Section 4.

### 3.1 Problem definition

Extractive QA datasets like SQuAD2.0 consist of triplets $(c_{src}, q_{src}, a_{src})$ of context paragraphs, questions, and answers, where[3]:

- *src* is the source language English, and
- $a_{src}$ is contained in $c_{src}$ (i.e., $a_{src} \subseteq c_{src}$); more precisely, $a_{src}$ is the specific substring of $c_{src}$ that answers question $q_{src}$.

Our goal is to generate the same triplets in the target language Basque, $(c_{tgt}, q_{tgt}, a_{tgt})$, where $a_{tgt} \subseteq c_{tgt}$ also holds. Conventional Machine Translation (MT) does not provide such a guarantee, whether due to the inclusion of spurious words not present in the context (e.g., articles), paraphrasing, or outright translation inaccuracies. In our study, compounding this issue is the fact that Basque is an agglutinative language with strong suffixing morphology, while English is primarily analytic by comparison. When translating an answer out of context, i.e., without considering the grammatical role it plays in the context it is found, the morphology of the phrase is bound to be altered, as illustrated in Figure 1. In brief, the main difficulty we address is rectifying these mismatches or corrupted instances where $a_{tgt} \nsubseteq c_{tgt}$.

### 3.2 Method

We propose a pipeline comprising the following steps to automatically generate the Basque version of SQuAD2.0: split, translate, align and compose.

**Sentence splitting** To ease the alignment step, we begin by splitting each SQuAD context $c_{en}$ into its individual constituent sentences $S_{en} = \{s_{en_1}, s_{en_2}, \ldots, s_{en_n}\}$. Then, questions and answers are mapped to the sentence the answer is found, obtaining new triplets $(s_{en_i}, q_{en}, a_{en})$, where $s_{en_i}$ denotes the $i$th sentence in $S_{en}$. We relied on a preexisting tool[4] to perform the segmentation.

**Machine Translation** We automatically translate this dataset into Basque, resulting in $(s_{eu_i}, q_{eu}, \tilde{a}_{eu})$ triplets. Here, $\tilde{a}_{eu}$ denotes the translated answer, which may or may not be found in the corresponding sentence $s_{eu_i}$ due to the nature of the translation process, as explained in Section 3.1.

[3]We deliberately omit *impossible* questions from this definition, as these do not pose any particular problem other than their translation itself.

[4]`github.com/mediacloud/sentence-splitter`

**Algorithm 1** Alignment of translated contexts and answers

```
for each context sentence s do
    N ← ngrams(s)
    E_N ← embed(N)
    for each question q and MT answers Ã of s do
        if q is not impossible then
            ã ← pick_longest(Ã)
            e_ã ← embed(ã)
            a ← arg min_N cosine(E_N, e_ã)
        end if
    end for
end for
```

This step was performed with Itzuli, a neural MT system accessible via API upon request[5]. The neural MT approach has demonstrated its robustness for Basque-Spanish translation (Etchegoyhen et al., 2018), and although performance on Basque-English was expected to be comparatively lower, considering scarcer training resources, the retrieved translations were considered of sufficient quality for the task. The Itzuli EN-EU system was trained on a mixture of parallel data available in the OPUS repository (Tiedemann, 2012) and synthetic data generated via back-translation (Sennrich, Haddow, and Birch, 2016).

**Answer alignment** At this stage, 61.98% and 60.04%, respectively, of the train and development translated answers $\tilde{a}_{eu}$ did not appear verbatim in the translated sentences $s_{eu_i}$. In other words, more than half the dataset was corrupted, as defined in Section 3.1. Therefore, our next step involves identifying the correct text span $a_{eu} \subseteq s_{eu_i}$ that is most similar to $\tilde{a}_{eu}$ and answers the question $q_{eu}$ for each $(s_{eu_i}, q_{eu}, \tilde{a}_{eu})$ triplet. To that end, we apply a semantically-oriented text similarity approach, drawing on contextual embeddings calculated from a neural language model. We make the assumption that, given a sufficiently robust language model, the embedding of the correct answer span $a_{eu}$ should be the closest to the embedding of the translated answer $\tilde{a}_{eu}$ for all possible spans of $s_{eu_i}$. The embedding of a text span is calculated as the mean pooling of all token embeddings within the span. The process is outlined in Algorithm 1[6] and is illustrated in Figure 2.

[5]`itzuli.vicomtech.org/api`

[6]In instances where multiple answers per question are present in the same sentence, which occurs when there are varying possible answers or criteria provided by different human annotators, we restrict the matching process to the longest answer. This approach is aimed at minimising instances where the answer con-

**Translated answer ($\tilde{a}_{eu}$)**

Text: X. eta XI. mendeak
Start: ??

| Rank | Sim. | Context sentence ($s_{eu_i}$) n-grams |
|---|---|---|
| ⭐ 1 | 0.778 | X. eta XI. mendeetan |
| 2 | 0.737 | eta XI. mendeetan |
| 3 | 0.702 | XI. mendeetan |
| 4 | 0.641 | eta XI. mendeetan Normandiari |
| 5 | 0.614 | eskualde bati, izena eman |
| 6 | 0.590 | eta XI. |
| 7 | 0.571 | bati, izena eman |
| 8 | 0.538 | XI. mendeetan Normandiari |
| 9 | 0.538 | X. eta XI. |
| 10 | 0.537 | izena eman zioten herriak |

**Aligned answer ($a_{eu}$)**

Text: X. eta XI. mendeetan
Start: 82

Figure 2: Top-10 most similar context n-grams to the translated answer of Figure 1. While they do not match exactly in wording and there are similar, potentially distracting candidates, the correct n-gram is assigned the highest score and is consequently chosen as the final answer span.

We have specifically experimented with the embeddings of two distinct models:

- **BERTeus** (Agerri et al., 2020) is a monolingual BERT-like language model trained for Basque. It adopts the conventional WordPiece tokenization approach (Schuster and Nakajima, 2012; Wu et al., 2016) used in foundational BERT models (Devlin et al., 2019), segmenting text spans into subwords. Our rationale for using BERTeus lies in the expectation that its better modelling of Basque language nuances, as compared to CANINE-s (next), would be an advantage in the semantic similarity task.

- **CANINE-s** (Clark et al., 2022) is also a Transformer-based language model. Unlike BERTeus, it is multilingual and operates at the character level, that is, it processes text by individual characters rather than tokenising it into subwords. Despite the lower amount of Basque content in CANINE-s' pretraining data compared to BERTeus, we have tested the former's utility based on the hypothesis that character-based tokenisation may be more robust to morphological variations of the same root word.

As a result of this step, we obtain well-formed $(s_{eu_i}, q_{eu}, a_{eu})$ triplets.

**Composition** In the last step, we reverse the sentence splitting process to reconstruct the original context-centred data structure. The sentences in $S_{eu} = \{s_{eu_1}, s_{eu_2}, \ldots, s_{eu_n}\}$ are collectively assembled to form the target context $c_{eu}$. Then, the $(s_{eu_i}, q_{eu}, a_{eu})$ triplets are straightforwardly mapped to $c_{eu}$, resulting in the desired $(c_{eu}, q_{eu}, a_{eu})$ triplets. Finally, basic post-processing is applied to remove any unwanted punctuation marks, such as trailing commas or parentheses, from the final aligned answers.

### 3.3 Result

**Size and format** Table 1 shows a quantitative overview of EusQuAD[7]. EuSQuAD contains as many contexts and questions as the original SQuAD2.0 dataset, and adheres to the same data structure (see an example in Figure 5). Therefore, any tool or script designed for processing SQuAD2.0 should be compatible with EuSQuAD. However, it must be noted that EuSQuAD differs from SQuAD2.0 in terms of the number of answers per question: unlike SQuAD2.0, which may feature multiple answers originating from different valid text segments or annotators, EuSQuAD includes only a single answer per question and sentence (refer to Footnote 6).

**Content** As EuSQuAD is a direct translation of SQuAD2.0, the distribution of question types in the former closely mirrors that of the latter, after accounting for linguistic variations —e.g., "what" can be translated as "zer" or "zein"; "zenbat" means both "how many" and "how much" (see Figures 6a and 6b). Furthermore, EuSQuAD is equally if not slightly more challenging than SQuAD2.0. We consider the following two statistics as proxies of the difficulty of each $(c, q, a)$ triplet: *a)* The overlap between contexts and questions; the greater the overlap, sists of only a short word or number, which could lead to incorrect matches due to heightened ambiguity. Our decision to prioritise the correctness of matches over coverage ensures greater accuracy.

[7] The statistics are the same regardless of the language model used to compute the alignments.

| | Train | Dev | Test |
|---|---|---|---|
| Contexts | 19,028 | 1,204 | 168 |
| Questions w/ answer | 86,734 | 5,921 | 490 |
| Impossible questions | 43,585 | 5,952 | - |
| Mean context length | 727 | 799 | 814 |
| Mean question length | 84 | 115 | 44 |
| Mean answer length | 22 | 26 | 23 |

Table 1: Statistics of EuSQuAD train and development sets, generated automatically from SQuAD2.0, and the new manually annotated test set. Mean text lengths are measured in characters.

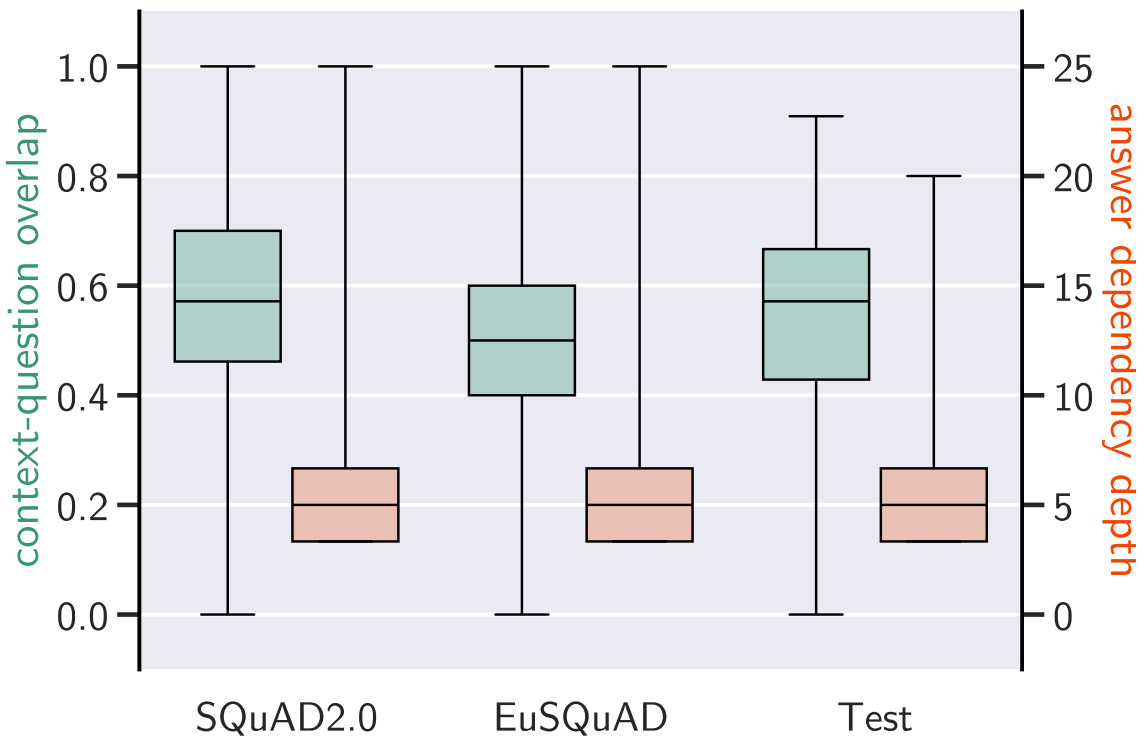

Figure 3: Overlap between contexts and questions (both lemmatised), in terms of ROUGE-L precision; and the maximum depth of the answers' dependency trees.

the easier it is to locate the answer. *b)* The depth of the dependency tree of the answers; the deeper the tree is, the more complex the answer is syntactically, therefore the more difficult it is to identify the exact answer span. As depicted in Figure 3, EuSQuAD contexts and questions exhibit slightly less overlap compared to SQuAD2.0, which can be seen as a favourable by-product of the translation process, while still maintaining comparable syntax complexity.

**Error analysis** Given the synthetic nature of EuSQuAD, an inherent degree of errors is to be expected. Thus, we manually analysed alignment inconsistencies in 200 question-answer pairs for each EuSQuAD variant, BERTeus and CANINE-s. The result of this analysis is shown in Figure 4. We observe that embeddings produced by CANINE-s are far better suited for the task at hand, having resulted in 77% correct alignments, as opposed to BERTeus' 50% (note that a baseline of literal matches yields an accuracy of 40%). Errors can be categorised into three types:

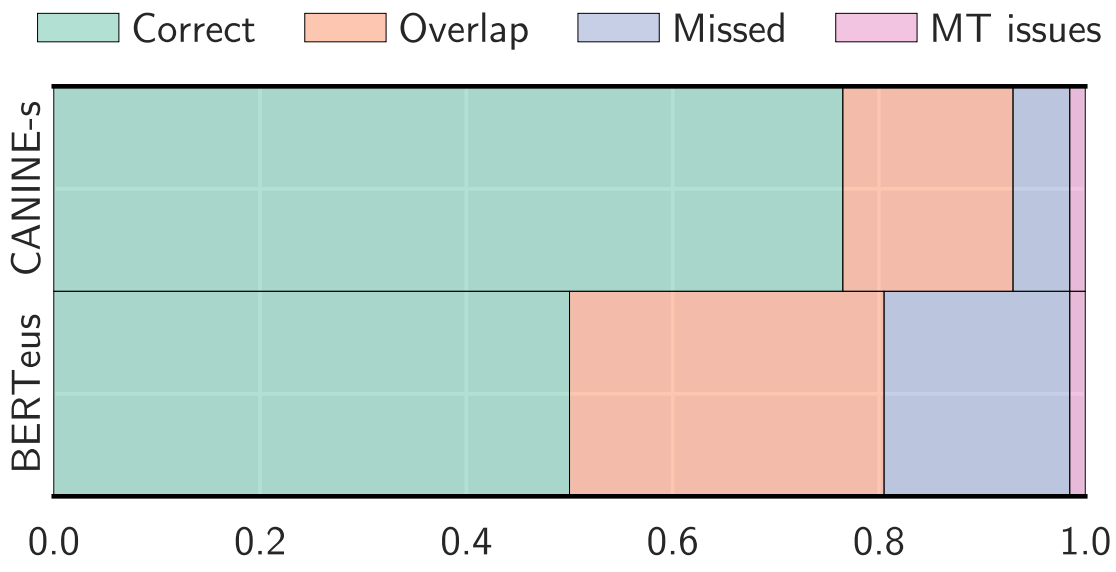

Figure 4: EuSQuAD alignment analysis

- The expected answer span and the correct span do not match exactly, but do overlap. In other words, the predicted span may include spurious content from adjacent phrases and/or fail to cover the expected answer completely. These comparatively less critical inconsistencies account for 17% and 30% of CANINE-s and BERTeus answers, respectively. In both cases, word-wise F1-score (as defined in Section 5.1) is approximately 75% when measured against manually corrected answer spans.
- The correct answer span is completely missed by CANINE-s and BERTeus in 6% and 18% of the analysed cases, respectively. In both cases, the most common distance between the predicted and the expected answer span is 0, i.e., the predicted answer is usually found immediately before or after the correct answer. Missed answers are likely to involve short spans or numbers, which are challenging to match with high confidence due to their increased ambiguity.
- Finally, only 3 errors are attributed to translation issues, where the correct answer could never be matched because it was omitted or incorrectly translated.

Appendix A presents examples of each error type and additional statistics from this manual analysis. All in all, our analysis reveals a significant reduction in alignment errors compared to prior datasets —refer to Section 2—, particularly with the CANINE-s variant. Moreover, given the prevailing notion that current QA models perform better when trained on large-scale noisy data rather than fewer high-quality examples (Carrino, Costa-jussà, and Fonollosa, 2020), we consider EuSQuAD to be a highly valuable contribution to the field.

```
{
 "qas": [
  {
   "question": "Zein herrialdetakoak ziren eskandinaviarrak
        ?",
   "id": "56ddde6b9a695914005b962a",
   "answers": [
    {
     "text": "Danimarka, Islandia eta Norvegiako",
     "answer_start": 247
    }
   ],
   "is_impossible": false
  },
  {
   "question": "Nor zen lider eskandinaviarra?",
   "id": "56ddde6b9a695914005b962b",
   "answers": [
    {
     "text": "Rollo",
     "answer_start": 306
    }
   ],
   "is_impossible": false
  },
  {
   "question": "Zein mendetan lortu zuten normandiarrek
        beren nortasun bereizia?",
   "id": "56ddde6b9a695914005b962c",
   "answers": [
    {
     "text": "X. mendearen lehen erdian",
     "answer_start": 628
    }
   ],
   "is_impossible": false
  },
  {
   "question": "Nork eman zion izena Normandiari 1000 eta
        1100 urteetan?",
   "id": "5ad39d53604f3c001a3fe8d1",
   "answers": [],
   "is_impossible": true
  },
  {
   "question": "Zertara dator Frantzia?",
   "id": "5ad39d53604f3c001a3fe8d2",
   "answers": [],
   "is_impossible": true
  },
  {
   "question": "Nori egin zion zin Karlos III.a erregeak?",
   "id": "5ad39d53604f3c001a3fe8d3",
   "answers": [],
   "is_impossible": true
  },
  {
   "question": "Noiz sortu zen identitate frankista?",
   "id": "5ad39d53604f3c001a3fe8d4",
   "answers": [],
   "is_impossible": true
  }
 ],
 "context": "Normandiarrak (normandieraz: Nourmands;
      frantsesez: Normandes; latinez: Normanni) X. eta XI.
       mendeetan Normandiari, Frantziako eskualde bati,
      izena eman zioten herriak izan ziren. Eskandinaviar
      jatorrikoak ziren (\"Norman\", \"Norseman\" etik
      dator), Danimarka, Islandia eta Norvegiako lapur eta
       piratak, eta, Rollo buru zutela, Karlos III.a
      Frantziakoa erregeari fedea zin egitea onartu zuten.
       Asimilazio belaunaldien bidez eta Frantziako eta
      Galiako jatorrizko populazioekin nahastuz, haien
      ondorengoek Mendebaldeko Frantziako karolingiar
      kulturekin bat egingo zuten pixkanaka. Normandiarren
       nortasun kultural eta etniko bereizia X. mendearen
      lehen erdian sortu zen hasieran, eta hurrengo
      mendeetan ere eboluzionatzen jarraitu zuen."
}
```

Figure 5: EuSQuAD has the same structure as SQuAD2.0, with context paragraphs and lists of questions, some of them *impossible* to answer with the given context.

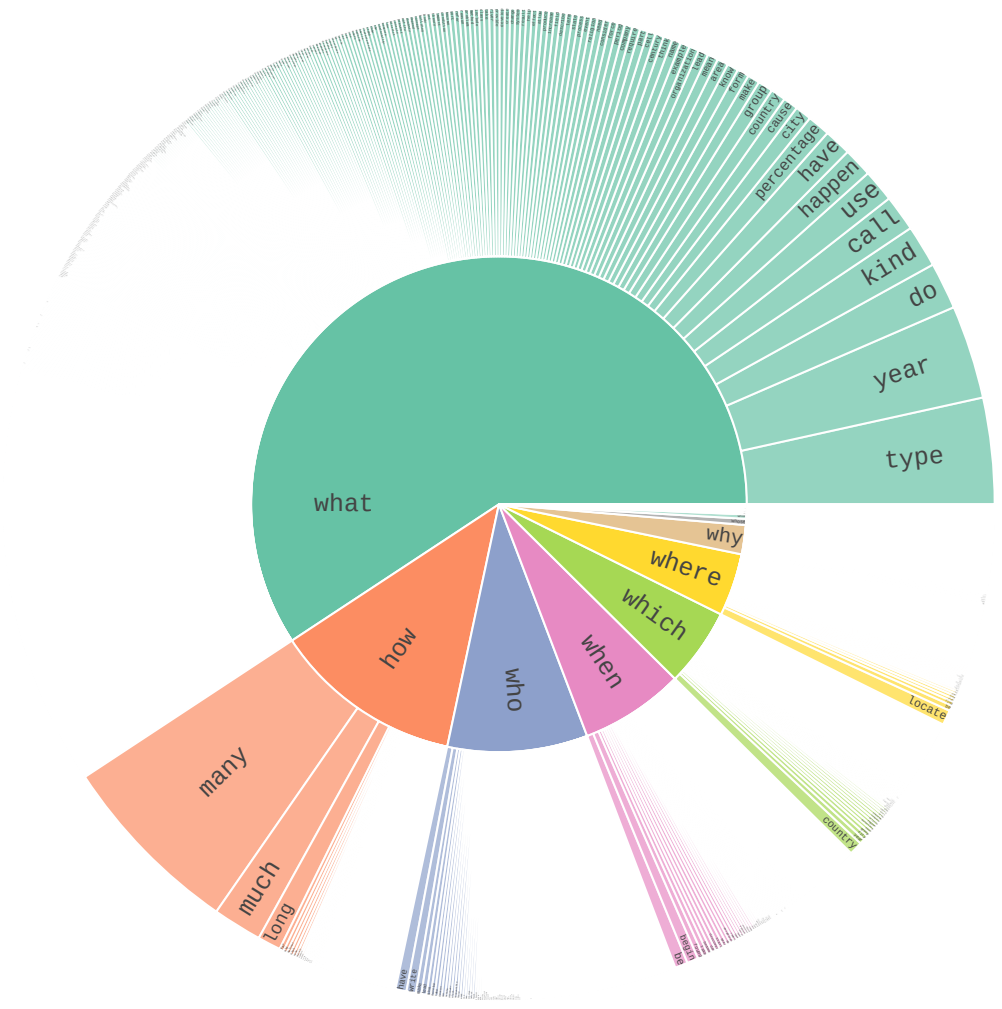

(a) SQuAD2.0 [dev]

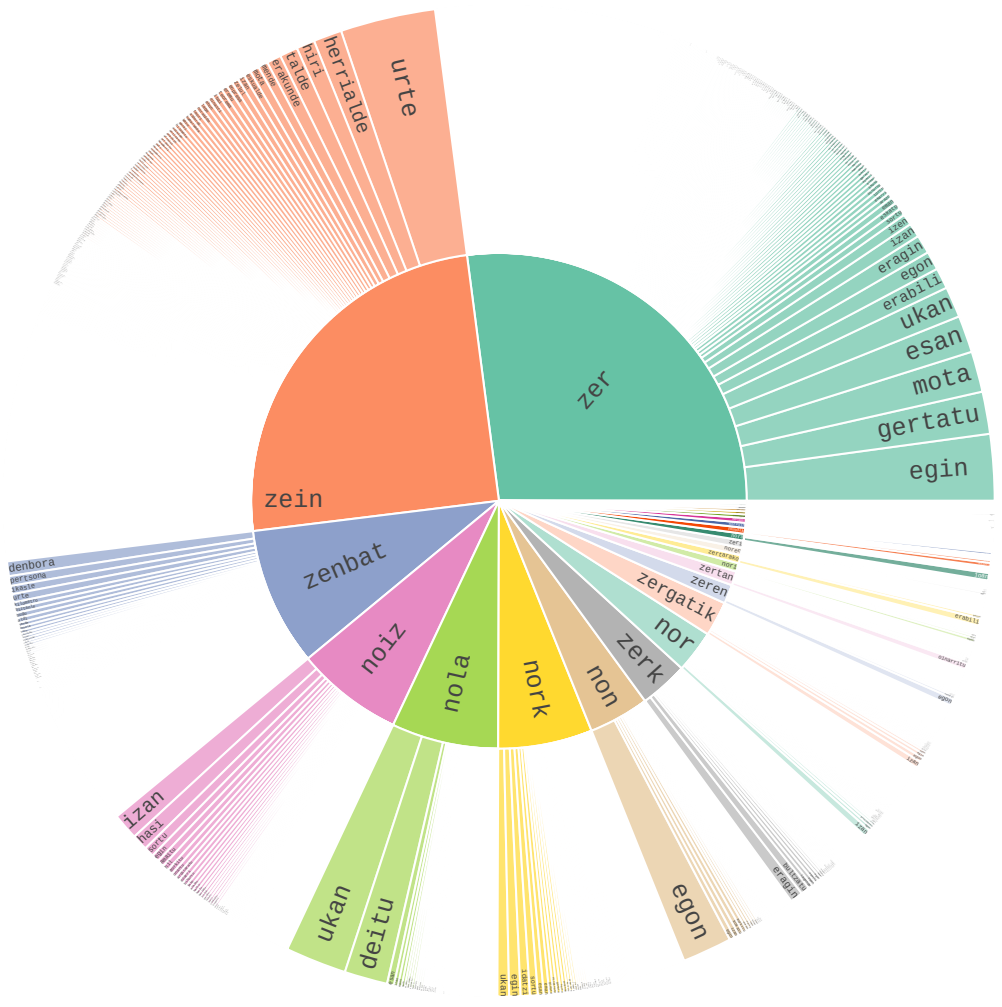

(b) EuSQuAD [dev]

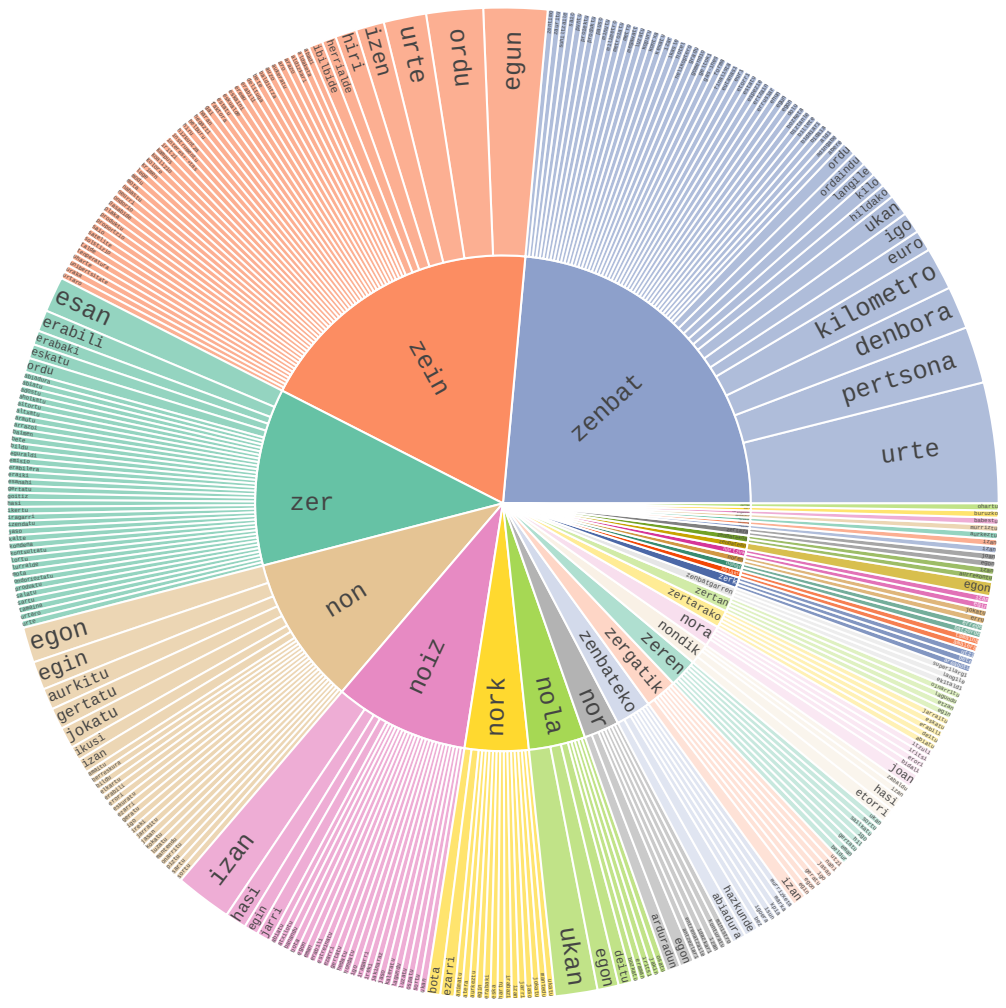

(c) Test

Figure 6: Interrogative words and their syntactic heads (lemmas).

## 4 *A new Basque QA test*

As EuSQuAD has been generated automatically and it does not include a test partition, we have manually labelled a new extractive QA dataset in order to be able to validate EuSQuAD through a set of experiments (in the next section, Section 5).

**Method** Our objective has been to create a quality and diverse extractive QA testing set in the style of SQuAD2.0. To that end, we collected context passages from Basque Wikipedia and local news articles. Using the Label Studio platform[8], one annotator read the articles, formulated questions, and annotated the corresponding answers. It must be noted that, in the present iteration, we did not pose *impossible* questions. An emphasis was placed on generating a wide variety of question types and avoiding literal quotations from the context. Subsequently, a second annotator reviewed the entire dataset for consistency and accuracy. Both annotators are native speakers of Basque.

**Result** The size of the new testing set is reported in Table 1, alongside EuSQuAD statistics. This test set comprises 490 question-answer pairs based on 168 context passages. Contexts and answers are comparable in length to those of EuSQuAD, although questions tend to be shorter. This should not be interpreted as making the test easier, as shorter questions may be less specific or more ambiguous. Additionally, the distribution of question types differs from that of EuSQuAD, with less imbalance towards "What" question types (cf. Figures 6b and 6c). Regarding the level of difficulty, the distribution of context-question overlap and answer dependency tree depth closely resembles that of SQuAD2.0, as shown in Figure 3.

## 5 *QA Experiments*

We conducted a series of QA experiments using EuSQuAD. The goal is twofold: on the one hand, to assess the value of EuSQuAD as a QA training dataset; on the other, to measure the downstream impact of the different language models used for alignment. As previously discussed (Section 3.3), CANINE-s —a multilingual, character language model— produced far more accurate answer alignments than BERTeus —a monolingual, subword-based model.

[8] labelstud.io/

| **Hyperparameter** | **Value** |
|---|---|
| Learning rate | 3E-5 |
| Maximum epochs | 10 |
| Early stopping patience | 5 |
| LR warmup epochs | 2 |
| LR schedule | Linear |
| Batch size | 4 |
| Gradient accumulation steps | 16 |
| Optimizer | AdamW |

Table 2: Fine-tuning hyperparameters

### 5.1 Experimental Setup

**Datasets** Our experiments involve four datasets. First, we trained three sets of QA models: one each for CANINE-s and BERTeus EuSQuAD variants, and another using SQuAD2.0. Second, our newly created dataset was employed as the experiments' testing set. That is, models trained on EuSQuAD were evaluated in a monolingual setting, while SQuAD2.0 models were evaluated in a cross-lingual setting.

**Models and training details** Following the current field standard, the assessed models consist in Transformer-based encoders trained for a span classification objective. The specific implementation can be found online.[9] As base models, we selected three medium-sized BERT-based language models:

- **Multilingual BERT** or **mBERT**[10] was pre-trained using the Wikipedias of more than 100 different languages. Although it can deal with numerous languages, it is not specialised in Basque.
- **IXAmBERT** (Otegi et al., 2020) was pre-trained on a more balanced mixture of Basque, Spanish, and English data.
- **BERTeus** (Agerri et al., 2020) is a monolingual model pre-trained exclusively on Basque data, as introduced earlier. This makes BERTeus one of the most suitable models —of this size and nature— to process Basque text.

Factoring in the three training datasets, we trained and evaluated a total of 9 different models. All of them were trained under the same settings and hyperparameters, as detailed in Table 2. Trainings continued until either reaching the maximum number of epochs, set at 10, or exhausting the early stopping patience of 5 epochs.

[9] huggingface.co/docs/transformers/model_doc/bert\#transformers.BertForQuestionAnswering

[10] github.com/google-research/bert/blob/master/multilingual.md

| Base model | Training set (variant) | F1 | EM |
|---|---|---|---|
| BERTeus | EuSQuAD (CANINE-s) | **65.7 ± 2.8** | **52.7 ± 3.0** |
| | EuSQuAD (BERTeus) | 59.3 ± 1.0 | 46.2 ± 1.1 |
| | SQuAD2.0 | 26.9 ± 2.0 | 20.7 ± 1.4 |
| IXAmBERT | EuSQuAD (CANINE-s) | **65.1 ± 3.3** | 51.3 ± 3.1 |
| | EuSQuAD (BERTeus) | 55.3 ± 2.4 | 42.9 ± 1.5 |
| | SQuAD2.0 | 62.8 ± 2.2 | **51.7 ± 1.2** |
| mBERT | EuSQuAD (CANINE-s) | **55.8 ± 3.5** | **44.6 ± 2.3** |
| | EuSQuAD (BERTeus) | 50.6 ± 4.4 | 39.0 ± 2.3 |
| | SQuAD2.0 | 36.2 ± 4.7 | 27.6 ± 4.2 |

Table 3: F1 and Exact Match (EM) scores on the manually labelled Basque test set. The best results for each base model are highlighted in **boldface**. Best overall results are underlined.

**Evaluation** Model performance is measured in terms of the two standard statistics for extractive QA, F1-score and Exact Match (EM), following established community recipes.[11] F1-score measures the number of overlapping words between the gold answer and the predicted answer, after removing punctuation and capitalisation. By *words* in this context we refer to whitespace-separated sequences of text. EM measures the number of predicted answers that are identical to the gold answers. Both metrics range between 0 and 100, a higher score denoting a better model performance.

## 5.2 Results

Table 3 shows the results of the evaluation. For each model, we computed the mean result and the standard deviation from three runs each using different random seeds, so as to mitigate biases arising from (un)lucky initial conditions.

The main result from these experiments is that models trained on EuSQuAD outperform those trained on SQuAD2.0, achieving the best scores overall in terms of both F1-score and exact matches. CANINE-s model variants performed consistently better than their BERTeus counterparts, under similar conditions. The previously noted 50% higher alignment error rate associated with BERTeus correlates with a downstream F1-score drop ranging from 5 to 16 points, depending on the base model. Interestingly, the optimal combination uses BERTeus as the base model. These results reinforce the potential superiority of character-based language models for span alignment purposes compared to token-based ones, even when the latter are more adept in the target language.

The performance of IXAmBERT trained on SQuAD2.0 is worth noting as well. First, it outperformed the use of all mBERT variants, even those fine-tuned on EuSQuAD. It also achieved the best absolute EM score among IXAmBERT variants, although by a small margin over the EuSQuAD CANINE-s variant. These results may be attributed to a better English-Basque transfer learning capability of the IXAmBERT model, which was jointly trained on a balanced mixture of English, Basque and Spanish data. Nonetheless, the model trained on EuSQuAD CANINE-s still achieved the highest F1-scores among IXAmBERT variants.

Overall, the results of these experiments demonstrate the usefulness of the EuSQuAD dataset for the task, in particular in its CANINE-s variant. Although there is still margin for improvement, considering the best absolute values in terms of F1-score and EM, our approach led to consistent gains across the board and provides a new baseline for the task of extractive QA in Basque.

## 6 *Discussion*

The development of EuSQuAD involved several design decisions with an impact on the final result. We have prioritised having a first version of the dataset as a baseline to support further experimentation and validation,

[11] hf.co/spaces/evaluate-metric/squad_v2

but the present work could be improved along several lines.

For instance, we opted to use a generic MT system to translate the dataset, although it might be worth exploring an adaptation of this module to the specifics of extractive QA. As noted in previous sections, translating the answer independently of the context produces mismatches that alter the quality of the resulting corpus. Extracting the answer directly from the translated context would eliminate this type of error, and could be performed via inline tags in the source context. This type of approach would require an adaptation of the MT system to support appropriate tag translation, which was outside the scope of this work.

Additionally, we decided to align only the longest translated answer per question and sentence. Most of the times, the answers attached to a single question are very similar to each other, but are located in different parts of the text. This contextual information is important for a QA model to achieve better generalisation. Future work may study the impact and the trade-off of aligning all the answers.

Further research is also needed to assess the correctness of the alignment and the different elements that impact the final result. About the CANINE-S versus BERTeus performance during the alignment step, we hypothesise that the flexibility of the character-space to represent the short answer spans outperforms the subword token-space. This topic deserves further study, especially since most existing pre-trained language models work with subword tokens.

Finally, the validity of our findings relies significantly on the quality of the manually created testing set, which, nonetheless, could also benefit from certain improvements. On the one hand, the incorporation of *impossible* questions would increase the difficulty level of the dataset, thereby offering a more realistic assessment of model robustness. On the other, expanding the dataset to include multiple valid answers per question would allow for a fairer evaluation and model comparison.

## 7 Conclusions

In this paper we have presented EuSQuAD, a version of SQuAD2.0 for Basque. Our approach is based on machine-translating the original corpus with a generic neural machine translation system, and addressing mismatches between context and answers via semantic text similarity. The resulting dataset is of the same size as the original SQuAD2.0 dataset (over 142k question-answer pairs), readily usable for QA-related tasks in Basque.[12]

To evaluate our approach, we compared the performance of several model variants trained on SQuAD or EuSQuAD against a new manually annotated dataset of 490 questions in Basque. The models trained on EuSQuAD achieved consistently better results than their SQuAD-trained counterparts.

We also examined the impact of using either character-based or token-based semantic text similarity approaches to generate EuSQuAD variants. Our results suggest that character-based language models can lead to significantly better performance on the downstream QA task, with a reduction in incorrectly aligned answer-spans. This finding opens a new avenue for further research.

## Acknowledgements

This work is partially funded by the Basque Business Development Agency, SPRI, under the project ADAPT-IA (KK-2023/00035) and by the Ministerio para la Transformación Digital y de la Función Pública and Plan de Recuperación, Transformación y Resiliencia - Funded by EU – NextGenerationEU within the framework of the project ILENIA with reference 2022/TL22/00215335. Our gratitude to the Vicomtech HSLT Department's Machine Translation team for providing the English-Basque translation service.

[12] `https://github.com/Vicomtech/EusQuAD`

## A *Alignment error analysis*

The manual analysis of 400 aligned answers (200 per alignment model, CANINE-s and BERTeus) unearthed three types of errors: inexact or overlapping answers; missed answers; and errors induced by the translation itself, due to the correct answer having been omitted or incorrectly translated. Figures 9 to 11 show examples for each of these error types. Having manually annotated the correct answers of these instances, we saw that the average F1-scores are 0.76 and 0.74 for CANINE-s and BERTeus, respectively, for overlap errors (see Figure 7). The distance, measured in words, between predicted and correct answer spans in missed alignments is shown in Figure 8.

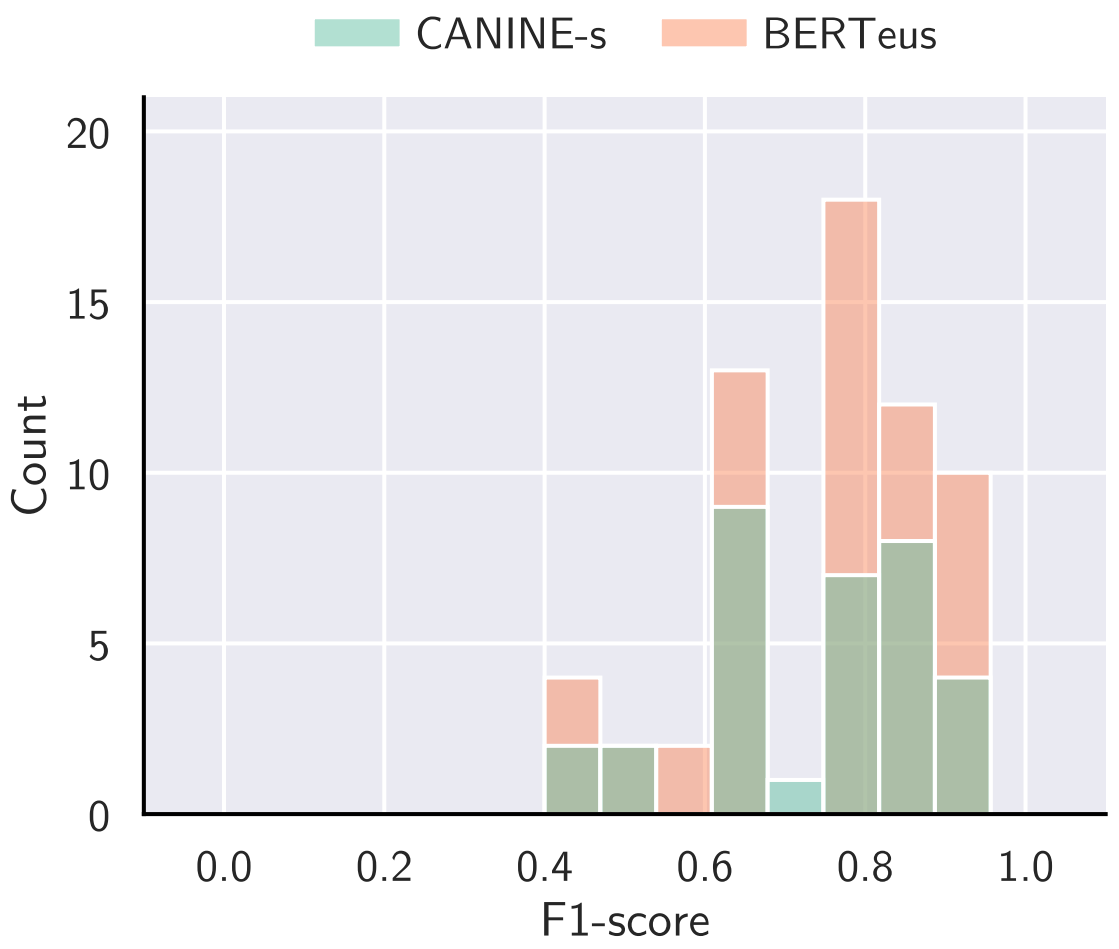

Figure 7: Distribution of F1-score for overlapping aligned answers.

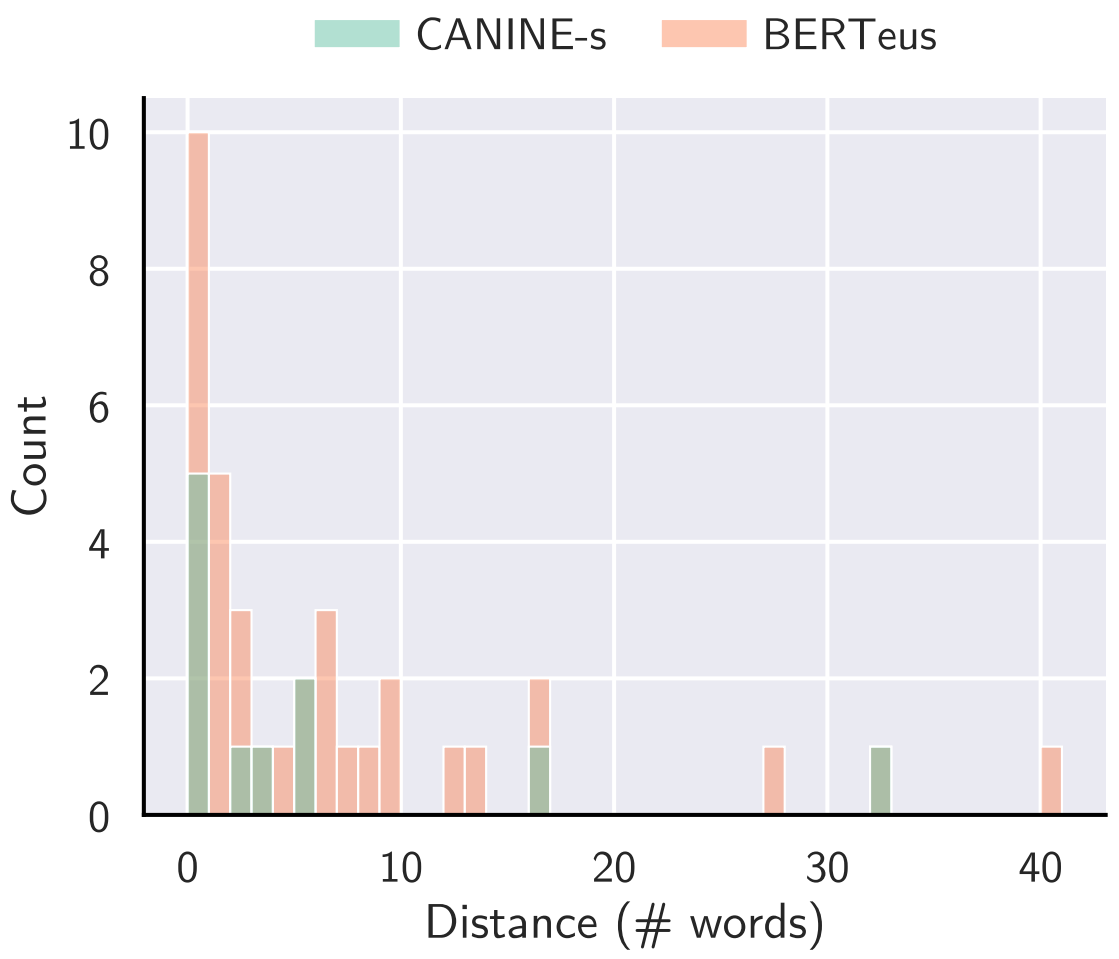

Figure 8: Distribution of the distance, in words, between predicted and correct spans.

**Question**

What are pharmacists forbidden to do?

**Answer in Context**

In most jurisdictions (such as the United States), pharmacists are regulated separately from physicians. These jurisdictions also usually specify that only pharmacists may supply scheduled pharmaceuticals to the public, and that pharmacists cannot form business partnerships with physicians or give them "kickback" payments. However, the American Medical Association (AMA) Code of Ethics provides that physicians may dispense drugs within their office practices as long as there is no patient exploitation and patients have the right to a written prescription that can be filled elsewhere. 7 to 10 percent of American physicians practices reportedly dispense drugs on their own.

**Question**

Zer debekatzen zaie farmazialariei?

**Answer in Context**

Jurisdikzio gehienetan (Estatu Batuetan, esaterako), botikariak medikuengandik bereizita arautzen dira. Jurisdikzio horiek zehazten dute, halaber, farmazialariek bakarrik eman ditzaketela programatutako farmazia-produktuak jendearentzat, eta farmazialariek ezin dutela negozio-aliantzarik egin medikuekin, eta ezin dietela "ordainketarik" eman. Hala ere, Amerikako Medikuen Elkartearen (AMA) Kode Deontologikoak xedatzen du medikuek sendagaiak eman ditzaketela beren bulego-jardunen barruan, pazientearen esplotaziorik ez dagoen bitartean eta pazienteek beste leku batean bete daitekeen errezeta idatzia jasotzeko eskubidea duten bitartean. AEBetako medikuen % 7 eta % 10 artean drogak beren kabuz ematen omen dituzte.

Figure 9: Example of an incorrectly aligned answer, where the predicted answer overlaps in part with the expected answer. This is the most common type of error among the analysed instances (70% and 61% of CANINE-s and BERTeus errors, respectively).

**Question**

Where did John Paul II celebrate Mass in Warsaw?

**Answer in Context**

John Paul II's visits to his native country in 1979 and 1983 brought support to the budding solidarity movement and encouraged the growing anti-communist fervor there. In 1979, less than a year after becoming pope, John Paul celebrated Mass in Victory Square in Warsaw and ended his sermon with a call to "renew the face" of Poland: Let Thy Spirit descend! Let Thy Spirit descend and renew the face of the land! This land! These words were very meaningful for the Polish citizens who understood them as the incentive for the democratic changes.

**Question**

Non ospatu zuen Joan Paulo II.ak meza Varsovian?

**Answer in Context**

Joan Paulo II.ak 1979an eta 1983an bere jaioterrira egindako bisitek laguntza eman zioten mugimendu solidarioari, eta han gero eta sutsuago zegoen antikomunista bultzatu zuten. 1979an, aita santu izan eta urtebete baino gutxiagora, Joan Paulok meza eman zuen Varsoviako Victory Squaren, eta Poloniaren "aurpegia berritzeko" deiarekin amaitu zuen sermoia: Jaitsi bedi zure izpiritua! Jaitsi bedi Zure Izpiritua eta berritu dezala lurraren aurpegia! Lur hau! Hitz horiek oso esanguratsuak ziren poloniar hiritarrentzat, aldaketa demokratikoen akuilu gisa ulertzen baitzituzten.

Figure 10: Example of an incorrectly aligned answer, where the predicted answer misses entirely the expected answer. This type of error accounts for 23% and 36% of CANINE-s and BERTeus errors in our analysis.

**Question**

How many counties initially made up the definition of southern California?

**Answer in Context**

In 1900, the Los Angeles Times defined southern California as including "the seven counties of Los Angeles, San Bernardino, Orange, Riverside, San Diego, Ventura and Santa Barbara." In 1999, the Times added a newer county—Imperial—to that list.

**Question**

Zenbat konderrik osatzen zuten hasieran Kaliforniako hegoaldeko definizioa?

**Answer in Context**

1900. urtean, Los Angeles Times egunkariak honela definitu zuen Kalifornia hegoaldea: "Los Angeles, San Bernardino, Orange, Riverside, San Diego, Ventura eta Santa Barbara konderriak." 1999an, Times egunkariak konderri berriago bat — Imperial — gehitu zuen zerrenda horretara.

Figure 11: Example of an incorrectly aligned answer induced by MT omission: the expected answer "zazpi" ("seven" in Basque) is not explicitly mentioned in the translated context. This type of error affects both models equally and occurs just 3 times on the analysed sample of 200 examples.